\documentclass[sigconf]{acmart}
\AtBeginDocument{%
  }

\setcopyright{rightsretained}
\copyrightyear{2025}
\acmYear{2025}
\acmDOI{3746252.3760959}
\acmConference[CIKM '25] {Proceedings of the 34th ACM International Conference on Information and Knowledge Management}{ November 10--14, 2025}{Seoul, Republic of Korea.}
\acmBooktitle{Proceedings of the 34th ACM International Conference on Information and Knowledge Management (CIKM '25), November 10--14, 2025, Seoul, Republic of Korea}
\acmISBN{979-8-4007-2040-6/2025/11}

\usepackage{xcolor}
\usepackage{listings}
\usepackage{kotex} 
\usepackage{graphicx}
\usepackage{booktabs}
\usepackage{multirow}
\usepackage{pifont}
\usepackage{subcaption} 
\usepackage{tabularx,array}  
\usepackage{makecell}
\newcolumntype{Y}{>{\centering\arraybackslash}X}  

\usepackage[inline]{enumitem}   
\usepackage{booktabs} 

\newcommand{\blue}[1]{#1}

\begin{document}

\title{\blue{Sparse and Dense Retrievers Learn Better Together:} \\ Joint Sparse-Dense Optimization for Text-Image Retrieval}


\author{Jonghyun Song}
\affiliation{%
  \institution{Graduate School of Data Science, Seoul National University}
  \city{Seoul}
  \country{South Korea}}
\email{hyeongoon11@snu.ac.kr}

\author{Youngjune Lee}
\affiliation{%
  \institution{NAVER Corporation}
  \city{Seongnam}
  \country{South Korea}}
\email{youngjune.lee93@navercorp.com}

\author{Gyu-Hwung Cho}
\affiliation{%
  \institution{NAVER Corporation}
  \city{Seongnam}
  \country{South Korea}}
\email{gyuhwung.cho@navercorp.com}

\author{Ilhyeon Song}
\affiliation{%
  \institution{NAVER Corporation}
  \city{Seongnam}
  \country{South Korea}}
\email{ilhyeon.song@navercorp.com}

\author{Saehun Kim}
\affiliation{%
  \institution{NAVER Corporation}
  \city{Seongnam}
  \country{South Korea}}
\email{saehun.kim@navercorp.com}

\author{Yohan Jo}
\authornote{Corresponding Author}
\affiliation{%
  \institution{Graduate School of Data Science, Seoul National University}
  \city{Seoul}
  \country{South Korea}}
\email{yohan.jo@snu.ac.kr}

%
 
\renewcommand{\shortauthors}{Jonghyun Song et al.}

\begin{abstract}
Vision-Language Pretrained (VLP) models have achieved impressive performance on multimodal tasks, including text-image retrieval, based on dense representations. Meanwhile, Learned Sparse Retrieval (LSR) has gained traction in text-only settings due to its interpretability and efficiency with fast term-based lookup via inverted indexes. \blue{Inspired by these advantages, recent work has extended LSR to the multimodal domain. However, these methods often rely on computationally expensive contrastive pre-training, or distillation from a frozen dense model, which limits the potential for mutual enhancement.} To address these limitations, we propose a simple yet effective framework that \blue{enables bi-directional learning between dense and sparse representations through Self-Knowledge Distillation}.
\blue{This bi-directional learning is achieved using an integrated similarity score—a weighted sum of dense and sparse similarities—which serves as a shared teacher signal for both representations. }
To ensure efficiency, we fine-tune the final layer of the dense encoder and the sparse projection head, enabling easy adaptation of any existing VLP model.
Experiments on MSCOCO and Flickr30k demonstrate that our sparse retriever \blue{not only outperforms existing sparse baselines, but also achieves performance comparable to—or even surpassing—its dense counterparts, while retaining the benefits of sparse models.}\footnote{Our code is available at \url{https://github.com/holi-lab/mm-sparse-retrieval}}

\end{abstract}




\maketitle

\section{Introduction}
The Vision-Language Pre-trained (VLP) models \blue{such as CLIP \cite{CLIP} and others ~\cite{blip, albef}} have significantly advanced multimodal research \blue{by demonstrating strong performance in tasks like image captioning ~\cite{captioning1, captioning2, captioning3}, visual question answering ~\cite{vqa1, vqa2}, and visual grounding ~\cite{grounding1, grounding2}.} These VLP models have also become widely adopted for cross-modal retrieval (i.e., text-to-image and image-to-text retrieval) ~\cite{retrieval1, retrieval2, retrieval3, retrieval4, retrieval5}, by projecting images and text into a shared dense embedding space. 

\blue{Meanwhile, in text retrieval, learned sparse retrieval ~\cite{splade, coil, deepct, sparse4} has emerged as a significant advancement.} In these methods, the sparse projection head converts dense representations from the pre-trained language models  ~\cite{bert} into sparse lexical representations over the vocabulary space.
Unlike dense representations, these sparse ones preserve explicit lexical features, making them easier to interpret and allowing the use of inverted-index infrastructure for fast term-based lookup ~\cite{bruch2024efficient, patil2011inverted}. As a result, they have emerged as a compelling research direction alongside dense retrieval ~\cite{lin2021few}.

Inspired by the success of sparse methods in text retrieval, recent studies have extended this idea to the cross-modal setting. Early approaches to \textit{learned sparse text-image retrieval} ~\cite{chen-etal-2023-stair, lexvla, lexlip} analogously transform the dense representations from VLP models into lexical representations with a sparse projection head. However, a key limitation of these approaches is their reliance on end-to-end pre-training. This process is computationally expensive because it requires training the entire model architecture from scratch on massive datasets, often consisting of millions of image-text pairs. 
This reliance on contrastive supervision also limits the expressiveness of the sparse retriever, as the learning signal is typically derived from coarse-grained binary relevance labels.
To address these limitations, distillation-based approaches have emerged, with D2S~\cite{d2s} replacing contrastive learning with a distillation objective \cite{splade-v2}. In this setting, the sparse projection head is trained to reproduce similarity scores generated by a frozen dense encoder.
\blue{However, the fixed nature of these dense representations imposes a key limitation in that a frozen dense encoder restricts the sparse projection head to expressing only the information already present in the dense embeddings.} Furthermore, although sparse representations could \blue{provide complementary signals to refine dense ones}, such feedback is rarely exploited. Most existing methods rely on unidirectional distillation—dense to sparse—thus underutilizing their potential synergy.

To overcome this challenge, we propose a simple yet highly effective framework \blue{for jointly optimizing dense and sparse representations through bi-directional self-knowledge distillation}. 
Unlike previous approaches, we distill knowledge from an integrated similarity score, which is computed as a weighted sum of dense and sparse similarities, \blue{to enable bi-directional distillation}. This integrated score serves as a teacher signal to supervise both the dense and sparse similarity score.
While adapting dense features is critical for optimizing their interaction with sparse objectives, fully fine-tuning VLP models typically demands substantial computational resources. To balance effectiveness and efficiency, we update only the sparse projection head and the final layer of the dense encoders (Fig.~\ref{fig:arch}). Despite its simplicity, our framework improves the retrieval performance of both sparse and dense retrievers. \blue{Experiments on MSCOCO and Flickr30k show that our framework not only enables our sparse retriever to achieve statistically significant improvements over state-of-the-art baselines, but also boosts the dense retriever's performance beyond its original backbone.}

\section{METHODS}

\subsection{Task Definitions}

\begin{figure}[t] 
\
\centering
\includegraphics[width=\linewidth]{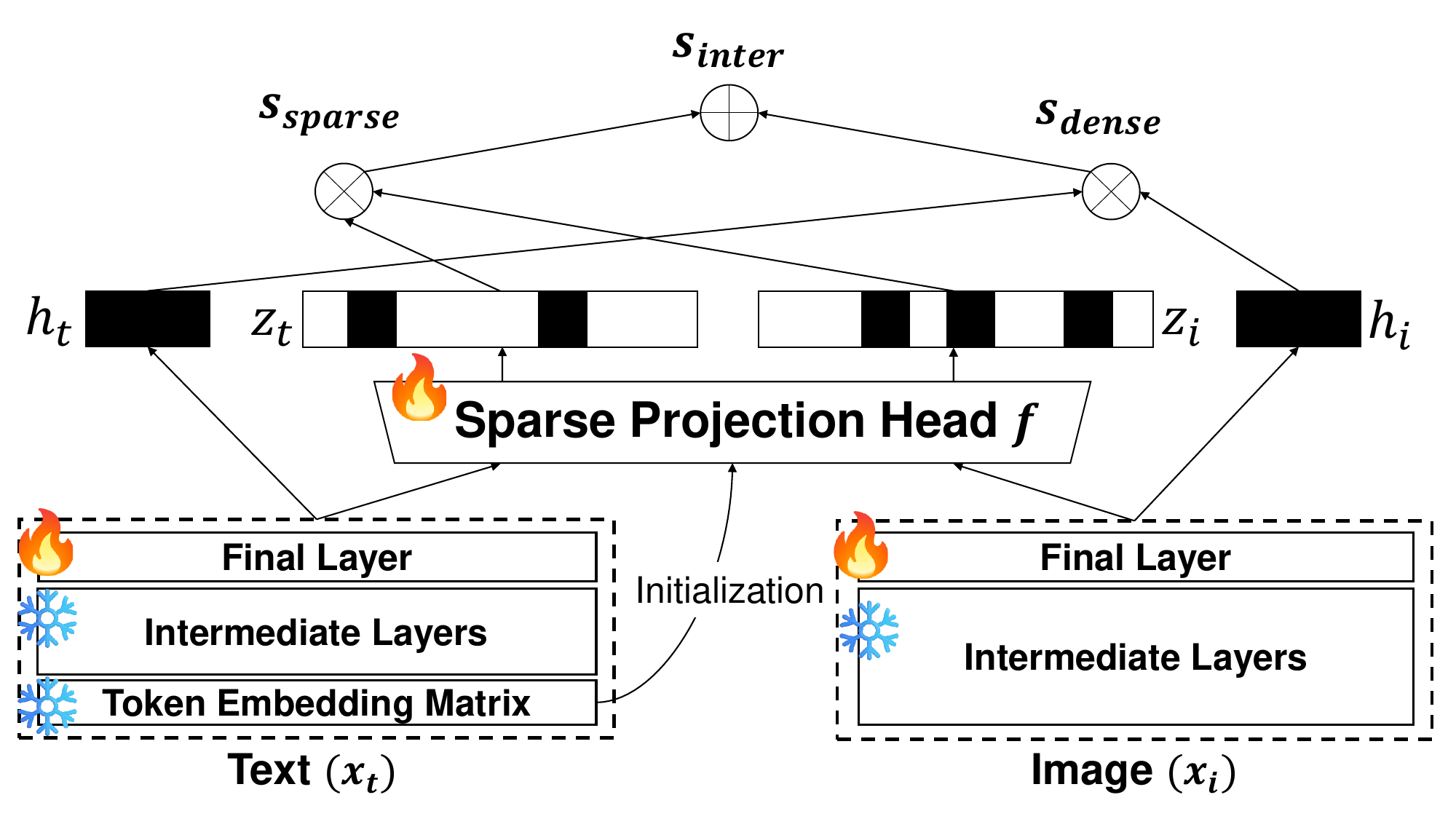} 
\vspace{-0.8cm}
\caption{Overall architecture. 
Text and image inputs are encoded by modality-specific encoders, with the final layer and the sparse projection head $f$ fine-tuned. This head maps the [CLS] embeddings ($h_t$, $h_i$) to sparse term weights ($z_t$, $z_i$).
}
\label{fig:arch}
\vspace{-0.3cm}
\end{figure}

In this work, we investigate learned sparse text-image retrieval~\cite{faghri2017vse} \blue{following the standard formulation ~\cite{d2s, blip, albef}.} Let $\mathcal{D} = \{ (x^{(p)}_i, \\ \{ x^{(p,q)}_t \}_{q=1}^k ) \}_{p=1}^N$ denote a multi-modal dataset consisting of $N$ image–text pairs, where \blue{each pair consists of an image $x^{(p)}_i$ and its \textit{k} associated captions  $\{ x^{(p,q)}_t \}_{q=1}^k$}. In this work, we focus on text-to-image retrieval, in which a textual caption $x_t$ is used as a query to retrieve the corresponding image $x_i$ from a large pool of candidates in $\mathcal{D}$. It directly mirrors real-world user behaviors in a wide range of visual content retrieval applications. Our goal is to develop a robust retriever that reliably ranks the correct image $x_i$ as the most relevant to the textual caption $x_t$.

\subsection{Architecture}
\label{ssec:arch}
We propose a novel framework for learned sparse text-image retrieval that jointly optimizes both dense and sparse representations to enable their mutual enhancement. The overall architecture, illustrated in Fig.~\ref{fig:arch}, begins by encoding a text $x_t$ and an image $x_i$ into dense \texttt{[CLS]} embeddings $h_t, h_i \in \mathbb{R}^d$, where $d$ is the embedding dimension.
$h_t$ and $h_i$ are transformed by the same sparse projection head $f: \mathbb{R}^d\rightarrow \mathbb{R}^{|V|}$ \blue{(where |V| denotes the vocabulary size)}, implemented as a simple multi-layer perceptron (MLP). It is then followed by ReLU and logarithmic transformations to compute the term importances $z_t$ , $z_i$ $\in \mathbb{R}^{|V|}$:

\begin{equation}
z_{*} = \log (1+\text{ReLU}(f(h_{*}))) \text{ where }* \in \{t, i\}
\end{equation}
\blue{To project the dense representations back into the vocabulary space, we initialize the projection head with the transposed word embedding matrix from the frozen text encoder, guiding it to produce semantically meaningful token-level scores.}

\blue{In terms of training, we enhance efficiency by freezing most layers of the backbone VLP encoder.} Specifically, we only fine-tune the final layer of the dense encoders and the sparse projection head. This strategy ensures that the model retains the strong generalization abilities of the pre-trained encoder, while efficiently adapting to sparse retrieval tasks with minimal training-time computation. 


\subsection{Self-Knowledge Distillation} 
\label{ssec:distill}
While sparse retrieval offers interpretability and efficiency, deploying it on top of dense VLP models poses a challenge: \blue{it requires learning new sparse representations while simultaneously preserving the rich semantic quality of the existing dense features.}
Preserving this quality is critical, as any degradation in the dense features will be directly inherited by the sparse representations that learn from them. Since the backbone encoders are originally pretrained with dense supervision, naively optimizing for sparse representations could degrade the underlying dense features. To address this, we propose a training strategy that jointly optimizes dense and sparse similarity scores, where the integrated score provides bidirectional supervision. In detail, the model is trained using a combination of contrastive loss and a self-knowledge distillation loss that jointly optimizes the dense and sparse scores.

To calculate these losses, we compute sparse and dense scores by taking the dot product of the text and image embeddings with a learnable temperature $\tau$: $s_{\text{dense}}(x_{t}, x_{i})  = \frac{\langle h_t , \, h_i\rangle}{\tau}$, $
s_{\text{sparse}}(x_{t}, x_{i}) = \langle z_t,\,z_i\rangle$. We also introduce an integrated score $s_{\text{inter}}=w_1 \cdot s_{\text{dense}} + w_2\cdot s_{\text{sparse}}$, which is a weighted sum of the individual score types. This integrated score acts as a soft teacher signal that dynamically reflects the complementary strengths of both representations. It supervises both dense and sparse scores in the self-knowledge distillation process, allowing them to learn from each other without losing their individual characteristics.

\paragraph{Contrastive Loss}
In both sparse and dense retrieval, we apply the InfoNCE  loss ~\cite{infonce}. The loss is computed over a batch of $N$ text-image pairs $(x_{t,m}, x_{i,m}) \ (1\leq m\leq N)$. For each score type $s_* \in \{s_{\text{dense}}, s_{\text{sparse}}, s_{\text{inter}}\}$, and for both directions (text-to-image and image-to-text), the bidirectional loss $\mathcal L^{s_*}_{a\rightarrow b} \ \text{where} \ (a, b)\in\{(t, i), (i,t)\}$ is defined as:

\begin{equation}
\mathcal L^{s_*}_{a\rightarrow b}
=-\frac{1}{N}\sum_{m=1}^N
\log
\frac{\exp\!\bigl(s_*(x_{a,m},x_{b,m})\bigl)}
{\displaystyle\sum_{j = 1}^N
\exp\!\bigl(s_*(x_{a,m},x_{b,j})\bigl)}
\label{eq:infonce}
\end{equation}

The final contrastive loss for each score type is then the average of these bidirectional losses: $\mathcal L_{s_*}
=\frac{\mathcal L^{s_*}_{i\rightarrow t}+\mathcal L^{s_*}_{t\rightarrow i}}{2}
\label{eq:contrastive}
$.

The overall contrastive loss is a weighted sum of these contrastive losses: $\mathcal{L} = \lambda_1 \cdot \mathcal{L}_{s_{\text{dense}}} + \lambda_2 \cdot \mathcal{L}_{s_{\text{sparse}}} + \lambda_3 \cdot \mathcal{L}_{s_{\text{inter}}}$, where $\lambda_1,\lambda_2$, and $\lambda_3$ are hyperparameters that balance each loss.

\paragraph{Distillation Loss}
To leverage the informative guidance of the integrated score ~$s_{\text{inter}}$, we adopt an auxiliary loss based on self-knowledge distillation. Inspired by recent advances in knowledge distillation ~\cite{chen2024bge, ensemble}, this method uses ~$s_{\text{inter}}$ as a teacher signal to supervise both the dense and sparse representations. 

Specifically, for a student score $s_* \in \{s_{\text{dense}}, s_{\text{sparse}}\}$, the distillation loss is defined as:
$
\mathcal{L}_{\text{distill}}(s_{\text{inter}}, s_*) = -p(s_{\text{inter}}) \cdot \log p(s_*)
$
where $p(\cdot)$ is the softmax function. This formulation is based on the cross-entropy loss and aims to align the probability distributions of the student and teacher scores.
The total distillation loss is then the average of the individual distillation losses for the dense and sparse components:
$
\mathcal{L}' = (\mathcal{L}_{\text{distill}}(s_{\text{inter}}, s_{\text{dense}}) + \mathcal{L}_{\text{distill}}(s_{\text{inter}}, s_{\text{sparse}}))/2
$

\paragraph{Overall Training Objective}
Finally, we combine contrastive and distillation loss with L1 sparsity regularization $L_1(\mathbf{x}) = |\mathbf{x}|$ with sparsity regularization parameters $\eta_t $ and $\eta_i$:
\begin{equation}
    \mathcal{L}_{final} = \mathcal{L}+\mathcal{L'}+\eta_tL_1(z_t)+\eta_iL_1(z_i)
    \label{eq: overall_loss}
\end{equation}
This holistic objective promotes a unified learning environment where the model benefits from both direct contrastive learning signals and knowledge distillation from the integrated score.

\section{Experiments}
\subsection{Experimental Setup}
\label{ssec:setup}

\paragraph{Data.}
We train and evaluate our models on the widely used MSCOCO~\cite{mscoco} and Flickr30k~\cite{flickr} datasets, using the standard Karpathy split~\cite{karpathy_split}. It splits each dataset into 113.2k/5k/5k (train/val/test) for MSCOCO and 29.8k/1k/1k for Flickr30k.

\paragraph{Implementation} 
\blue{Following prior work \cite{d2s}, we implement our models by loading BLIP\cite{blip} and ALBEF\cite{albef} checkpoints that have already been fine-tuned on each dataset.}
In training, we train the sparse projection head along with the final layer of the text and image encoders. 
All models, including ours and re-implemented baselines, are tuned using the same set of hyperparameters \blue{including learning rates, regularization weights ($\eta_t$ and $\eta_i$), and loss balancing parameters ($\lambda_{1,2,3}$ and $w_{1,2}$).} We train models for 200 epochs and quadratically increase the regularization loss weights $\eta_t$ and $\eta_i$ (Eq.~\ref{eq: overall_loss}). Models are trained using PyTorch Lightning, and training takes approximately four hours on a single RTX 3090 GPU (24GB VRAM).



\paragraph{Metrics \& Evaluation} We report Recall@$\{1, 5\}$ and MRR@10 for test sets. For sparse and dense results, we rank the images according to the sparse score $s_{\text{sparse}}$ and the dense score $s_{\text{dense}}$, respectively.  
\blue{We compare our approach against sparse (VisualSparta \cite{visualsparta} and LexLIP \cite{lexlip}) and dense (BLIP \cite{blip} and ALBEF \cite{albef}) baselines, using results reported in their original papers. For a more direct comparison, we re-implemented D2S and trained it under the same conditions as our own model.}
\\
\subsection{Results}
\label{ssec:main_experiment}
\newcommand{\midgap}{\hspace{0.8em}}

\begin{table}[]
\caption{Sparse text-to-image retrieval on MSCOCO and Flickr30k. Bold indicates the best runs. \blue{\textsuperscript{\dag} denotes a statistically significant improvement of our model over the corresponding baseline (paired t-test, p < 0.05).}}
\centering
\begin{tabularx}{\linewidth}{l Y Y Y @{\midgap} Y Y Y}
\hline
\textbf{Model} & \multicolumn{3}{c}{\textbf{MSCOCO}} & \multicolumn{3}{c}{\textbf{Flickr30k}} \\
               & \small R@1 & \small R@5 &  \small M@10 & \small R@1 & \small  R@5 & \small M@10 \\ \hline
VisualSparta   & 45.1 & 73.0 & -     & 57.1 & 82.6 & -     \\
LexLIP         & 53.2 & 79.1 & -     & 78.4 & 94.6 & -     \\
D2S (ALBEF)    & 51.9 & 78.9 & 63.4  & 77.3 & 94.3 & 84.6  \\
D2S (BLIP)     & 55.9 & 81.1 & 66.7  & 81.3 & 95.9 & 87.6  \\\hline
Ours (ALBEF)   & 53.2$^{\dag}$ & 79.3$^{\dag}$ & 64.3$^{\dag}$  & 78.6$^{\dag}$ & 95.1$^{\dag}$ & 85.8$^{\dag}$  \\
Ours (BLIP)    & \textbf{57.6}$^{\dag}$ & \textbf{82.4}$^{\dag}$ & \textbf{68.1}$^{\dag}$ & \textbf{82.0}$^{\dag}$ & \textbf{96.4}$^{\dag}$ & \textbf{88.2}$^{\dag}$ \\ \hline
\end{tabularx}
\label{tab:main-result}
\vspace{-0.3cm}
\end{table}


\paragraph{Comparison with Sparse Retrieval Baselines} We report the sparse retrieval results in Table ~\ref{tab:main-result}. Our model, fine-tuned from BLIP, consistently outperforms all other sparse baselines across both datasets and evaluation metrics. This highlights the superior retrieval effectiveness of our proposed method.
In detail, comparing D2S and Ours (both fine-tuned from BLIP and ALBEF) demonstrates that our strategy is more effective when trained from the same vision-text encoders.
These findings imply that the bi-directional supervision from our self-knowledge distillation leads to better sparse representations, an advantage not captured by standard unidirectional distillation methods.

\begin{table}[]
\caption{Evaluation of dense and our models on MSCOCO and Flickr30k. ALBEF and BLIP refer to the backbone model performance. \textit{Ours (Sparse)} and \textit{Ours (Dense)} denote retrieval performance with $s_{\text{sparse}}$ and $s_{\text{dense}}$, respectively. }
\centering
\begin{tabular*}{\linewidth}{@{\extracolsep{\fill}}lcccccc}
\hline
\textbf{Model} & \multicolumn{3}{c}{\textbf{MSCOCO}} & \multicolumn{3}{c}{\textbf{Flickr30k}} \\
               & \small R@1 & \small R@5 &  \small M@10 & \small R@1 & \small  R@5 & \small M@10 \\ \hline
\multicolumn{7}{l}{\textit{ALBEF-based}} \\
ALBEF          & 53.3 & 79.4 & 64.3 & 79.1 & 94.9 & 86.6 \\
Ours (Sparse)  & 53.2 & 79.3 & 64.3  & 78.6 & 95.1 & 85.8 \\
Ours (Dense)   & 54.5 & 80.5 & 65.6 & 79.7 & 95.4 & 86.6 \\ \hline
\multicolumn{7}{l}{\textit{BLIP-based}} \\
BLIP           & 57.0 & 82.0 & 67.8 & 83.2 & 96.7 & 89.3 \\
Ours (Sparse)  & 57.6 & 82.4 & 68.1 & 82.0 & 96.4 & 88.2 \\
Ours (Dense)   & 58.7 & 82.9 & 69.1 & 82.8 & 96.7 & 88.7 \\ \hline
\end{tabular*}
\label{tab:dense-result}
\vspace{-0.3cm}
\end{table}




\paragraph{Comparison with Dense Retrieval Baselines} 
We compare our model against dense baselines (ALBEF and BLIP) in Table~\ref{tab:dense-result}. Notably, our sparse retriever (\textit{Ours (Sparse)}) often surpasses the dense counterparts, despite offering significantly greater efficiency and interpretability. Moreover, when applied to dense retrieval (\textit{Ours (Dense)}), it also improves performance over the original baselines. \blue{These results demonstrate that our method effectively transforms knowledge from pretrained dense encoders into sparse representations, while preserving competitive performance of dense models.}


\subsection{Analysis}
\label{ssec:analysis}

\paragraph{Ablation Study} We conduct an ablation study using the ALBEF backbone to investigate the individual effects of our proposed components, with results presented in Table ~\ref{tab:ablation1}. For the best configuration of each dataset, we evaluate two primary ablation settings: (1) \blue{Self-Knowledge Distillation}: We analyze the impact of our self-knowledge distillation loss by training the model solely with the contrastive loss (i.e., removing $\mathcal{L}'$ from Eq. \ref{eq: overall_loss}). (2) \blue{Final Layer Fine-tuning}: We compare models that fine-tune only the final layer of the VLP encoder against those that freeze the entire encoder. 

Each result highlights the individual and combined contributions of the proposed techniques. (1) Comparing the full model (Row 3: \ding{51} \ding{51}) with the variant that replaces self-knowledge distillation by contrastive loss (Row 2: \ding{55} \ding{51}) reveals the benefit of our distillation objective. On both MSCOCO and Flickr30k, self-distillation leads to a consistent improvement in retrieval accuracy over the contrastive-only baseline, suggesting that joint dense-sparse optimization helps the sparse representations better capture high-quality semantics than learning them in isolation.
(2) We analyze the effect of fine-tuning the encoder’s final layer. Compared to the variant that keeps the encoder frozen (Row 1: \ding{51} \ding{55}), fine-tuning the final layer (Row 3: \ding{51} \ding{51}) yields stronger performance. This indicates that even minimal update of the encoder contributes meaningfully to sparse representation learning, whereas freezing the final layers restricts mutual learning between dense and sparse representations, as the dense embeddings remain fixed.

\setlength{\tabcolsep}{2pt} 
\begin{table}[]
\caption{Ablation study on the final layer fine-tuning and self-knowledge distillation. \ding{51} \ding{51} is our best configuration.}
\label{tab:ablation1}
\resizebox{\linewidth}{!}{%
\begin{tabular}{ccllllll}
\hline
 
&\multicolumn{1}{l}{}                                                   & \multicolumn{3}{c}{\textbf{MSCOCO}} & \multicolumn{3}{c}{\textbf{Flickr30k}} \\
 \begin{tabular}[c]{@{}c@{}}  Self-Knowledge\\   Distillation\end{tabular} 
&\begin{tabular}[c]{@{}c@{}}\ Final Layer \\ Fine-tuning \end{tabular}& R@1       & R@5       & M@10       & R@1        & R@5        & M@10       \\ \hline
 \ding{51}                                                             
& \ding{55}            & 52.1& 78.6& 63.4& 76.8& 94.8&84.7\\
 \ding{55}            &\ding{51}&           51.9&           78.9&              63.4&            77.5&            94.8&              85.0\\
 \ding{51}                                                             &\ding{51}            &           53.2 & 79.3 & 64.3  & 78.6 & 95.1 & 85.8\\ \hline
\end{tabular}%
}
\vspace{-0.3cm}
\end{table}

\paragraph{Effectiveness-Efficiency Trade-off}

\begin{figure}[t]
  \centering

\includegraphics[width=\linewidth]{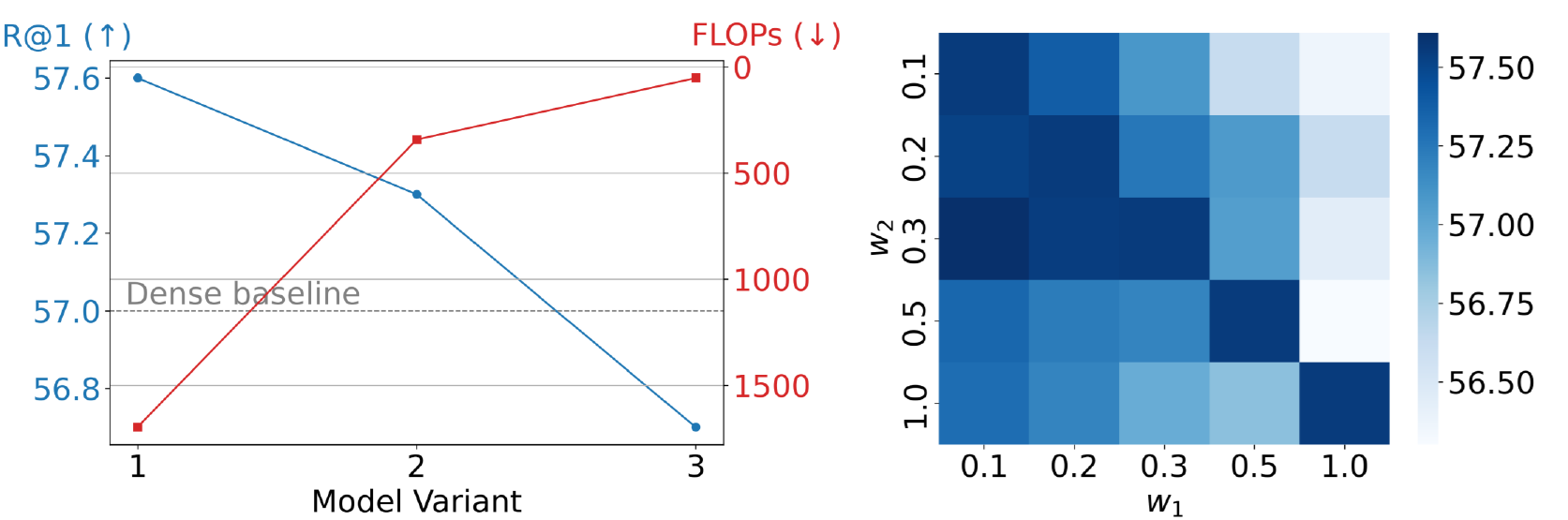}
\vspace{-0.5cm}
\caption{(\textit{Left}) Effectiveness vs. efficiency trade-off. Model variant 3 applies Probabilistic Expansion Control (PEC), whereas variants 1 and 2 do not. \textit{(Right)} R@1 performance heatmap under varying dense ($w_1$) and sparse ($w_2$) score weights.}
    \label{fig:flops}
    \vspace{-0.5cm}
\end{figure}

The trade-off between effectiveness and efficiency is a critical consideration in sparse retrieval. To improve efficiency, previous work~\cite{d2s} introduced Probabilistic Expansion Control (PEC), a method designed to prevent sparse representations from over-activating irrelevant terms. Figure~\ref{fig:flops} (Left) examines the impact of PEC on this trade-off between retrieval effectiveness and computational efficiency, based on experiments run with BLIP on MSCOCO. We measure computational efficiency using FLOPs, \blue{defined as the average number of operations required for a similarity computation between a text and an image.}\footnote{This metric is formally defined as the expectation $\mathbb{E}_{t,i}(\sum_{j\in V}p_j^{t}p_j^{i})$, where $p_j^{t}$ and $p_j^{i}$ denote the activation probabilities for token $j$ in a given text $t$ and image $i$.}. 
\blue{Variants 1 and 2 represent earlier and later checkpoints from the same training run, both trained without PEC. They achieve better retrieval accuracy than the dense backbone model (the gray dashed line), albeit with higher computational cost. In contrast,}
Variant 3, with PEC, exhibits substantially lower FLOPs while maintaining competitive R@1 performance.  These results highlight a practical trade-off: depending on application constraints, one may choose a more accurate model at the expense of efficiency (Variant 1), or a more efficient model with slightly reduced accuracy (Variant 3). 

\paragraph{Dense-Sparse Weighting}
In Figure~\ref{fig:flops} (Right), we analyze the impact of varying the dense ($w_1$) and sparse ($w_2$) weights in the integrated score ($s_{inter} = w_1\cdot s_{dense}+w_2\cdot s_{sparse}$). The heatmap shows that assigning a higher weight to the sparse score ($w_2$) generally leads to better R@1 performance, especially when $w_1$ remains moderate (e.g., 0.1–0.3). In contrast, large $w_1$ values tend to degrade performance. These results suggest that sparse scores can serve as effective learning signals in their own right—an aspect that has been largely overlooked in prior work, which depends on dense-to-sparse distillation.

\section{CONCLUSION}
In this work, we propose a sparse representation learning framework for text-image retrieval.
Our approach leverages self-knowledge distillation through an integrated similarity score \blue{to jointly optimize both dense and sparse representations.}
 Our sparse retriever achieves state-of-the-art performance on standard cross-modal retrieval benchmarks, often surpassing fully fine-tuned dense models. \blue{Furthermore, the same training strategy also improves the performance of the dense models beyond their original backbones, demonstrating the general applicability of our framework.} These results highlight the potential of lightweight, alignment-aware techniques for \blue{building scalable and effective text-image retrieval systems.}

\noindent
\section*{Acknowledgement}
This work was supported by the National Research Foundation of Korea (NRF) grant funded by the Korean government (MSIT) (No. RS-2024-00414981) and by the National Supercomputing Center with supercomputing resources including technical support (KSC-2024-CRE-0554).

\newpage
\section*{GenAI Usage Disclosure}
This work utilized Generative AI solely for proofreading and editing purposes.
\bibliographystyle{ACM-Reference-Format}
\bibliography{sample-base}
\end{document}